\documentclass{article}
\usepackage{ijcai19}

\usepackage{times}
\usepackage{tikz}
\usepackage{soul}
\usepackage{url}
\usepackage[hidelinks]{hyperref}
\usepackage{hyperref}
\usepackage[small]{caption}
\usepackage{graphicx}
\usepackage{multirow}
\usepackage{amsmath}
\usepackage{booktabs}
\usepackage{algorithm}
\usepackage{hhline}
\usepackage{algorithmic}
\usepackage{CJKutf8}

\usetikzlibrary{positioning,arrows,shapes,shapes.multipart,shapes.geometric,fit}
\pgfdeclarelayer{back}
\pgfsetlayers{back,main}

\title{Improving Multilingual Sentence Embedding using Bi-directional Dual Encoder with Additive Margin Softmax}

\author{
Yinfei Yang
\and
Gustavo Hernandez Abrego\and
Steve Yuan\and
Mandy Guo\and
Qinlan Shen\and \\
Daniel Cer\and
Yun-hsuan Sung\and
Brian Strope\and
Ray Kurzweil
\affiliations
Google AI Language\\
\emails
\{yinfeiy, gustavoha\}@google.com
}

\begin{document}

\maketitle

\begin{abstract}
In this paper, we present an approach to learn multilingual sentence embeddings using a bi-directional dual-encoder with additive margin softmax. The embeddings are able to achieve state-of-the-art results on the United Nations (UN) parallel corpus retrieval task. In all the languages tested, the system achieves P@1 of 86\% or higher. We use pairs retrieved by our approach to train NMT models that achieve similar performance to models trained on gold pairs. We explore simple document-level embeddings constructed by averaging our sentence embeddings. On the UN document-level retrieval task, document embeddings achieve around 97\% on P@1 for all experimented language pairs. Lastly, we evaluate the proposed model on the BUCC mining task. The learned embeddings with raw cosine similarity scores achieve competitive results compared to current state-of-the-art models, and with a second-stage scorer we achieve a new state-of-the-art level on this task.

\end{abstract}

\section{Introduction}
\label{sec:intro}
Neural machine translation (NMT) systems are highly sensitive to the volume and quality of the parallel data, known as bitext, used for model training.
This motivates increased interest in collecting large amounts of parallel data corpora where filtering is used to guarantee quality.
\cite{jakob2010,antonova2011} have shown that it is possible to mine parallel documents from the internet using large distributed systems,
with computationally intensive and heavily engineered subsystems.
Recently, lightweight end-to-end word and sentence embedding-based approaches have gained popularity and are showing some success for this purpose ~\cite{gregoire2017,BOUAMOR18,schwenk2018filtering,mandy2018,artetxe2018a}.
These systems are usually easier to train as they require very little feature engineering.

One popular type of approach is based on cross-lingual embeddings, 
where two sentences in a translation pair are set to be close to each other in an embedding space.
In this approach, given a source sentence, nearest-neighbor search can be used in the cross-lingual embedding space to find potential translation candidates.
The nearest neighbors can be chosen in terms of cosine distance.
However, current approaches produce noisy matches that require a re-scoring step in order to obtain a clean parallel corpus for training an NMT system~\cite{mandy2018,artetxe2018a}.

\begin{figure}
  \centering
  \includegraphics[width=.43\textwidth]{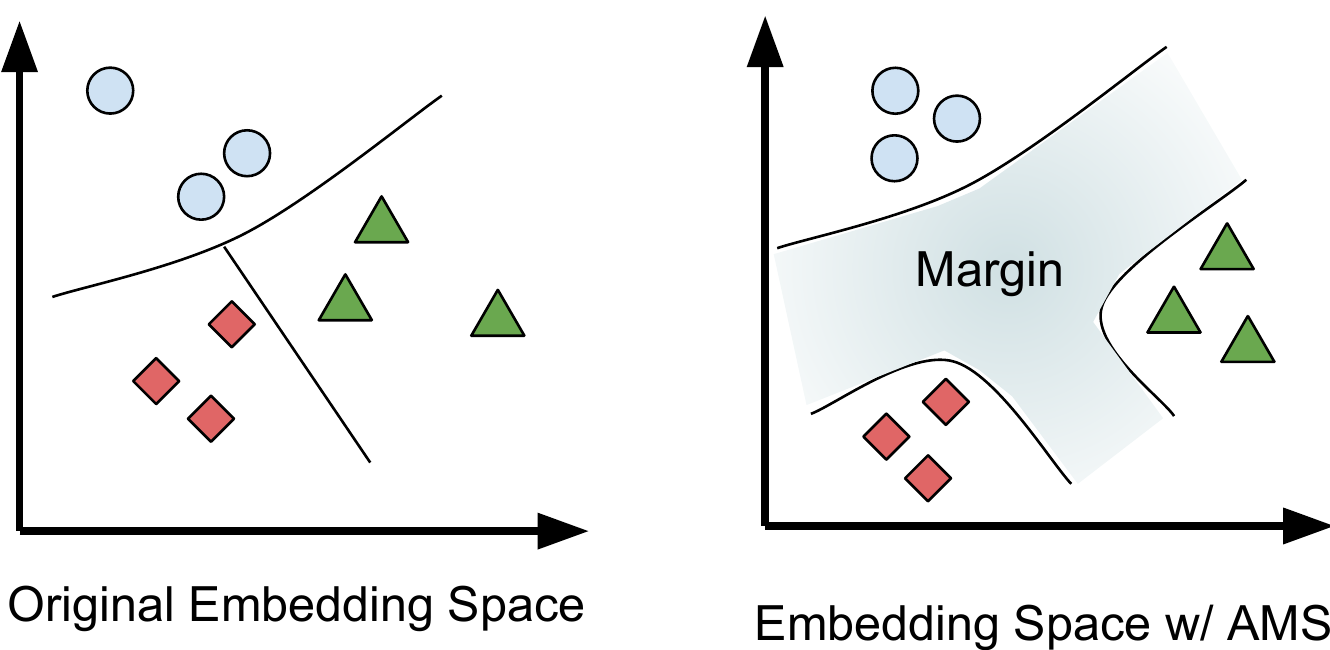}
  \caption{Embedding space with and without additive margin softmax. Shapes represent sentences that are translations of each other in different languages.
  An additive margin improves the separation between translations and nearby non-translations.
  }
  \label{fig:ams}
\end{figure}


We explore using a bi-directional dual encoder with additive margin softmax~\cite{ams}. The intuition is that we want to maximally separate sentences that are true translations of each other from similar sentences with overlapping but not identical meanings. Introducing a margin into the softmax attempts to achieve a fixed distance between class members and their respective decision boundaries. As shown in Figure \ref{fig:ams}, the margin around the decision boundaries tends to structure the embedding space such that examples from the same class have a more compact organization.


Extensive experiments demonstrate the learned multilingual sentence embeddings achieve state-of-the-art parallel sentence retrieval results.
On the United Nations (UN) Parallel Corpus~\cite{uncorpus}, the system achieves P@1 of 86\% or higher (from 11.3 million candidates) in all the languages tested.
NMT models trained on the mined parallel data are found to not only match or exceed the performance of both models trained on more complex filtering strategies but are also competitive with models trained on gold data. When the approach is extended to document-level embeddings by simply averaging all sentence embeddings, we
achieve around 97\% on P@1 (from 86k candidates) on the UN document-level retrieval task.

On the BUCC bitext mining task~\cite{bucc2018},
using our sentence embeddings achieves competitive results compared to state-of-the-art models, with F-scores ranging from 84.6 to 89.2. These results significantly outperform the baseline models by around 10 points.
With a second stage scorer we achieve a new state-of-the-art level, with F-scores ranging from 93.38 (ru-en) to 97.24 (de-en).

We summarize our contributions as follows:
\begin{itemize}
    \item[i] We introduce a novel bidirectional and additive margin softmax into the dual encoder framework for the bitext retrieval task;
    \item[ii] Empirical results show that the proposed approach greatly stabilizes cosine similarity in the multilingual embedding space, which also leads to state-of-the-art results on the UN sentence-level bitext retrieval task; 
    \item[iii] We also introduce an approach to embed documents by simply averaging all sentence embeddings, which achieves nearly state-of-the-art results on the UN document-level bitext retrieval task;
    \item[iv] We apply sentence level retrieval together with final ranking scoring to improve the state of the art on the BUCC mining task.
\end{itemize}

\section{Related Work}
Obtaining high-quality parallel corpora is one of the most critical problems in machine translation.
One longstanding approach for extracting parallel corpora is to mine documents from the web \cite{resnik1999mining}. 
Early approaches leveraged metadata such as document publication dates, titles or document structure \cite{yang2002mining,munteanu2005improving,munteanu2006extracting,chen2000parallel,resnik2003web,shi2006dom}. 
However, these approaches suffered from the fact that metadata associated with documents can be sparse or unreliable \cite{jakob2010}.

Another line of work consists of trying to identify bitexts using only textual information.
Some text-based approaches for this task rely on methods such as n-gram scoring \cite{jakob2010}, named entity matching \cite{do2009mining}, and cross-language information retrieval \cite{utiyama2003reliable,munteanu2005improving}.
One common characteristic of these approaches is that they require very heavy feature and system engineering.

Recent work on using embedding-based approaches has shown promising.
Texts are mapped to an embedding space in order to determine whether they are bitexts. 
\cite{gregoire2017} use a Siamese network \cite{yin2015abcnn} to map source and target language sentences into a shared embedding space. Then, a classifier is built to decide whether the sentences are parallel.
\cite{hassan2018achieving,schwenk2018filtering,artetxe2018a} formulate mulitilingual sentence embeddings in a shared space using the encoder states from a shared-encoder NMT system.
The cosine similarity between these sentence embeddings is used as a measure of cross-lingual similarity.

\cite{mandy2018} proposed a new approach using a dual-encoder architecture instead of a encoder-decoder one. The dual-encoder architecture optimizes the cosine similarity between the source and target sentences directly. Here, we extend this approach by using a bidirectional dual-encoder with additive margin softmax, which significantly improves the model performance.

\section{Model}
\label{sec:model}
This section introduces our bidirectional dual-encoder model for bitext mining trained with additive margin softmax.

\subsection{Dual Encoder Model}
\label{sec:de}

\begin{figure}
  \centering
  \includegraphics[width=.25\textwidth]{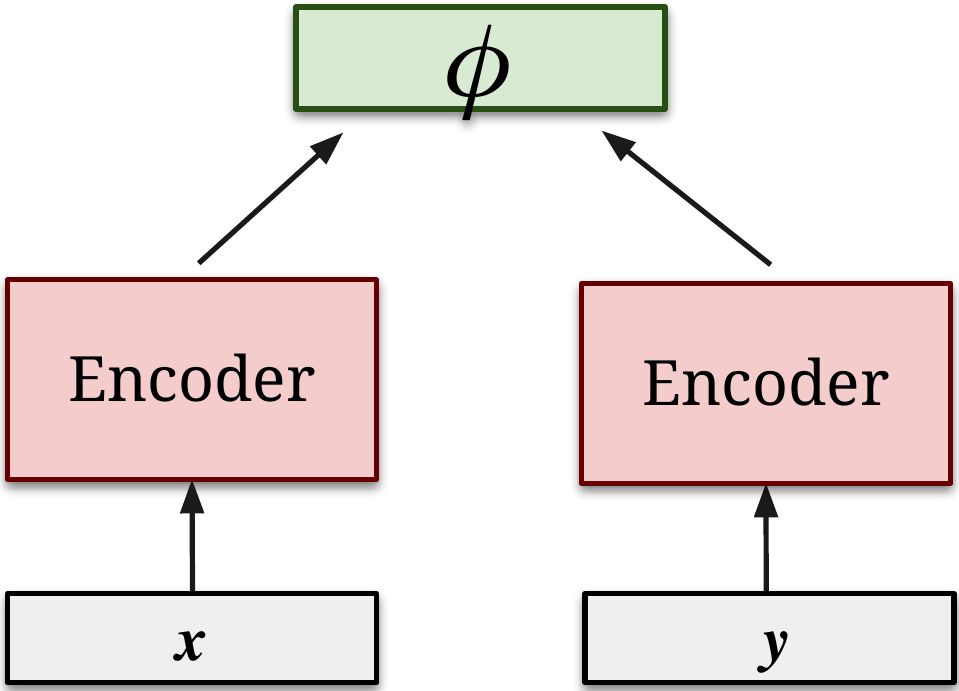}
  \caption{Dual encoder architecture.}
  \label{fig:dual_encoder}
\end{figure}

Figure \ref{fig:dual_encoder} illustrates a dual-encoder model, consisting of paired encoders feeding a combination function. Such models are well suited for ranking problems such as conversation response prediction and translation candidate prediction~\cite{yang2018,mandy2018}. Following \cite{mandy2018}, translation retrieval can be modeled as a ranking problem to place $y_i$, the true translation of $x_i$, over all other sentences in $\mathcal{Y}$.
$P(y_i \mid x_i)$ can be expressed as the following log-linear model, with $\phi$ scoring the compatibility between the encoded representations of $x_i$ and $y_i$:

\begin{equation}
\label{eq:bayes}
P(y_i \mid x_i) = \frac{e^{\phi(x_i, y_i)}} {\sum_{\bar{y} \in \mathcal{Y}} e^{\phi(x_i, \bar{y})}}
\end{equation}

$P(y_i\! \mid\! x_i)$ is approximated during training by sampling negatives, $y_n$, among translation pairs in the same batch:

\begin{equation}
\label{eq:approx}
P_{approx}(y_i \mid x_i) = \frac{e^{\phi(x_i, y_i)}} {e^{\phi(x_i, y_i)} + \sum_{n=1, n \neq i}^{N} e^{\phi(x_i, y_n)}}
\end{equation}

When using dot-product as the scoring function, $\phi$, a single matrix multiplication can be used to efficiently compute scores for all examples in the same batch. For source and target pairs, $x_i$ and $y_i$, the model can be optimized using the log-likelihood objective:  

\begin{equation}
\label{eq:softmax}
\mathcal{L}_s = -\frac{1}{N} {\sum_{i=1}^{N}\log \frac{e^{\phi(x_i, y_i)}} { e^{\phi(x_i, y_i)} + \sum_{n=1, n \neq i}^{N} e^{\phi(x_i, y_n)}}}
\end{equation}

\subsection{Bidirectional Dual Encoder}

$\mathcal{L}_s$ attempts to identify the correct $y_i$ for each $x_i$. Good embedding space representations of $y_i$ and $x_i$ should in principle allow retrieval in both directions, i.e. from both $x_i$ to $y_i$ and $y_i$ to $x_i$. We introduce a bidirectional learning objective, $\bar{\mathcal{L}}_s$ that explicitly optimizes both forward and backward ranking:

\begin{equation}
\label{eq:softmax_backward}
\mathcal{L}_s' = -\frac{1}{N} {\sum_{i=1}^{N}\frac{e^{\phi(y_i, x_i)}} { e^{\phi(y_i, x_i)} + \sum_{n=1, n \neq i}^{N} e^{\phi(y_i, x_n)}}}
\end{equation}

\begin{equation}
\label{eq:softmax_combined}
    \bar{\mathcal{L}}_s = \mathcal{L}_s + \mathcal{L}_s'
\end{equation}

\subsection{Dual Encoder with Additive Margin Softmax}
  
Additive margin softmax extends the scoring function $\phi$ by introducing a large margin, $m$, around positive pairs~\cite{ams}.

\begin{equation}
    \label{eq:scoring_am}
    \phi'(x_i, y_j) =
        \begin{cases}
          \phi(x_i, y_j) - m    & \quad \text{if } i = j \\
          \phi(x_i, y_j)        & \quad \text{if } i \neq j
        \end{cases}
\end{equation}

The margin, $m$, improves the separation between translations and nearby non-translations. Using $\phi'(x_i, y_j)$ with the bidirectional loss $\bar{\mathcal{L}}_s$, we obtain the additive margin bidirectional loss, $\mathcal{L}_{ams}$.

\begin{equation}
\label{eq:softmax_ams}
\mathcal{L}_{ams} = -\frac{1}{N} {\sum_{i=1}^{N}\frac{e^{\phi(x_i, y_i) - m}} { e^{\phi(x_i, y_i) - m} + \sum_{n=1, n \neq i}^{N} e^{\phi(x_i, y_n)}}}
\end{equation}

\begin{equation}
\label{eq:softmax_ams_combined}
    \bar{\mathcal{L}}_{ams} = \mathcal{L}_{ams} + \mathcal{L}_{ams}'
\end{equation}

\begin{table*}[!ht]
\small
\centering
    \begin{tabular}{l | r r r || r r r || r r r || r r r} 
        \hline
        \multirow{2}{*}{Models} & \multicolumn{3}{c||}{en-fr} & \multicolumn{3}{c||}{en-es} & \multicolumn{3}{c||}{en-ru} & \multicolumn{3}{c}{en-zh} \\
        \cline{2-13}
        &  P@1 & P@3 & P@10  &  P@1 & P@3 & P@10 &  P@1 & P@3 & P@10  &  P@1 & P@3 & P@10\\ 
        \hline
        \cite{mandy2018}          & 48.9 & 62.3 & 73.0 & 54.9 & 67.8 & 78.1 & - & - & - & - & - & -\\
        \cite{artetxe2018a}  & 83.3 & - & - & 85.8 & - & - & - & - & - & - & - & -\\ \hline
        DE          & 80.7 & 87.9 & 91.2 & 85.6 & 92.7 & 95.1 & 83.9 & 89.7 & 92.0 & 82.2 & 91.1 & 94.1 \\
        BiDE        & 82.3 & 90.7 & 94.2 & 86.3 & 93.0 & 95.6 & 85.7 & 92.3 & 95.1 & 83.1 & 91.3 & 94.6 \\
        BiDE+AM     & {\bf 86.1} & {\bf 93.5} & {\bf 96.1} & {\bf 89.0} & {\bf 95.2} & {\bf 97.2} & {\bf 89.2} & {\bf 94.8} & {\bf 96.9} & {\bf 87.9} & {\bf 94.7} & {\bf 97.1} \\
        \hline
    \end{tabular}
\caption{Precision at N (P@N) (\%) of target sentence retrieval on the UN corpus. Models attempt to select the true translation target for a source sentence from the entire corpus (11.3 million aligned sentence pairs). \protect\cite{mandy2018} is a dual encoder (DE) model trained with deep averaging network (DAN) instead of transformer. In the last row, AM stands for ``Additive Margin softmax".}
\label{tab:target_UN}    
\end{table*}

\section{Experimental Setup}
In this section, we describe the training data and provide details about the training process.

\subsection{Training Data}
The training corpus is extracted from the internet using a bitext mining system similar to the approach described in \cite{jakob2010}.
The extracted sentence pairs are filtered by a pre-trained data-selection scoring model~\cite{wei2018}.
Human annotators manually evaluate sentence pairs from a small subset of the harvested pairs and mark the pairs as either GOOD or BAD translations.
The data-selection scoring model threshold is chosen such that 80\% of the retrained pairs from the manual evaluation are rated as GOOD.
We select around 400 million sentences pairs for en-fr, en-es, en-de, en-ru and en-zh.\footnote{We started with a large dataset.  However, preliminary results showed that we could use only 10\% of the data without downgrading the model performance in the hold-out dev set. }
For each language pair, we use $90\%$ of the sentence pairs for training and $10\%$ as development set for parameter tuning.
We evaluate the trained models on the United Nations Parallel Corpus reconstruction task, and the BUCC bitext mining task.

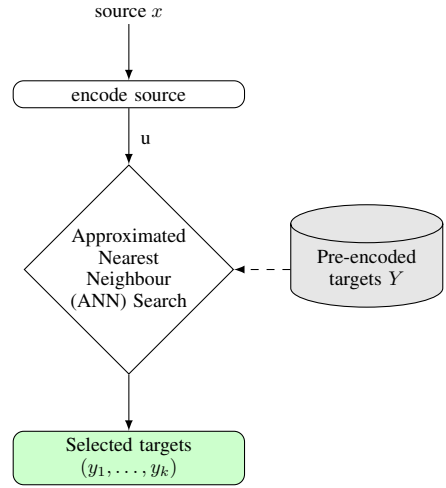
\begin{figure}
  \centering
  \begin{tikzpicture}[scale=0.75, every node/.style={transform shape}]

  \tikzstyle{arrow}=[->, thin, -latex]
  \tikzstyle{decision}=[diamond, draw=black, text width=6em, text centered, inner sep=1pt]
  \tikzstyle{box}=[rectangle, draw=black, text width=11em, text centered, rounded corners]
  \tikzstyle{dbase}=[cylinder, draw=black, text width=7em, text centered, shape border rotate=90, aspect=0.25, fill=black!10]

  \node[] (input) {source $x$};
  \node[box, below=of input] (encoder) {encode source};
  \node[decision, below=of encoder] (annsearch) {Approximated Nearest Neighbour (ANN) Search};
  \node[dbase, right=of annsearch] (targetset) {Pre-encoded \mbox{targets} $Y$ };
  \node[box, fill=green!20, below=of annsearch] (finaltargets) {\mbox{Selected targets} $(y_1, \ldots , y_k)$};

  \path
    [arrow] (input) edge (encoder)
    [arrow] (annsearch) edge (finaltargets)
  ;
  \path [arrow] (encoder) edge node [near start, xshift=3mm, yshift=-3mm] {u} (annsearch);
  \path
    [arrow, dashed] (targetset) edge  (annsearch)
  ;
\end{tikzpicture}
  \caption{
  \small{Translation retrieval pipeline, a source/target could be a sentence or document. Figure from \protect\cite{mandy2018}.}
  }
 \label{fig:retrieval_pipeline}
\end{figure}

\subsection{Model Configuration}
Our models merge word-level and character-level input representations.
For each language pair, we build a shared 200k unigram vocabulary, augmented with 10k out-of-vocabulary (OOV) hash buckets, that each map to 300 dim.\ word embeddings. For character-level representations, we decompose each token into all character $n$-grams with a length within the range $N=[a,b]$.\footnote{$N$=[1, 4] for en-zh and $N$=[3, 6] for other language pairs. We use different settings of $N$ for the en-zh pair because collocations of more than 3 character tokens are rare in Chinese.} The character $n$-grams are hashed to one of 200k hash buckets that map to 300 dim.\ embeddings. For each word, we then sum the embeddings for its character $n$-grams with its word-level embedding to arrive at the final token representation. 

Within a language pair, the encoding weights are shared for both languages.
The encoder uses a 3-layer transformer network architecture~\cite{vaswani2017attention}.
In the transformer layers, we use 8 attentions heads, a hidden size of 512, and a filter size of 2048.
Across the final transformer layer, we make use of 4 pooling layers consisting of max pooling, mean pooling, first token pooling, and attention pooling~\cite{attentive_pooling}. The 4 pooling layers are concatenated and then projected to a fixed 500-dim.\ embedding space.\footnote{During training, a vector length penalty is applied to the sentence embeddings. The length penalty is realized as the $\ell2$ norm of an embedding and is weighted within the loss by the constant 11.}.

A margin value of 0.3 is used in all experiments.
Training uses SGD with a 0.003 learning rate and batch size of 100. The learning rate is reduced to 0.0003 after 33 millions steps. Training concludes at 40 million steps. To improve training speed, we follow \cite{mutty2018} and multiply the gradients of the word and character embeddings by a factor of 25. We follow \cite{mandy2018} and append 5 additional hard negatives for each example. Parameters are tuned on the development set.

\subsection{The Parallel Corpus Retrieval Pipeline}

As mentioned in section \ref{sec:de}, the source embedding and target embedding can be encoded separately using a dual-encoder model.
Taking advantage of this fact, we use approximate nearest neighbor (ANN) search~\cite{ann} to find the candidate targets for each given source, once all target embeddings are built, and vice versa.
Figure \ref{fig:retrieval_pipeline} illustrates the retrieval pipeline. 
Source and target can be either sentences or documents in the following experiments.

\begin{table}
\small
    \centering
    \begin{tabular}{| l | r r r r|}
    \hline
     & en-fr & en-es & en-ru & en-zh \\ \hline
    Forward  & 86.1 & 89.0 & 89.2 & 87.9 \\
    Backward & {\bf 88.4} & {\bf 90.6} & {\bf 90.4} & {\bf 88.9} \\ \hline
    \end{tabular}
    \caption{P@1 (\%) on the UN corpus from forward search and backward search. Forward search treats English as source and the other language as target. Backward is vice versa.}
    \label{tab:target_UN_forward_backward}
\end{table}

\section{Evaluation on the United Nations Corpus}

The United Nations Parallel Corpus~\cite{uncorpus} contains 86,000 bilingual document pairs in five language pairs: from en to fr, es, ru, ar and zh.
Each document pair is also aligned at sentence level.
For each language pair, there are a total of 11.3 million aligned sentence pairs.\footnote{About 9.5 million unique translations remain for each language pair after de-duping.}

\subsection{Sentence-Level Retrieval}
We first apply the proposed model to retrieve target candidates at sentence-level from the entire UN corpus using approximated nearest neighbor search~\cite{ann}.
Precision at N (P@N) is used as evaluation metric for target retrieval, for N=1, 3, 10. We compare the performance with two baseline models.
The first baseline is a dual encoder model~\cite{mandy2018} trained with deep averaging network (DAN) and separate word embeddings for each language.
The second is from \cite{artetxe2018a} and extracts the encoders from  encoder-decoder NMT models. 

Table \ref{tab:target_UN} shows the precision of nearest neighbor search using en sentences to retrieving translations in fr, es, ru and zh.
The baseline systems are listed in the first two rows. 
Rows 3 to 5 show the results from our models: a unidirectional dual-encoder (DE) model, a bi-directional dual-encoder (BiDE) model, and the bi-directional model with additive margin softmax (BiDE+AM).
The BiDE is slightly but consistently better than DE for all four language pairs. 
BiDE+AM further improves performance over BiDE, with a minimum of 86.1\% P@1 in all language pairs. BiDE+AM greatly outperforms the baseline models in both en-fr and en-es, and establishes a new state-of-the-art in performance for this task.

We report the backward search performance in table \ref{tab:target_UN_forward_backward}, whereby we use each non-English sentence to search for its English translation.
Interestingly, the P@1 from the backward search is higher for all four language pairs. This suggests the embedded sentence representations achieve better discrimination over sentences with related but not identical meanings for English than for other languages.

\begin{table}
\small
    \centering
    \begin{tabular}{|l|r|r|}
        \hline
                        &  \textbf{en-fr} & \textbf{en-es} \\
                        & \textbf{(wmt14)} & \textbf{(wmt13)} \\ \hline
        Oracle          & 30.96  &  28.81 \\
        \cite{mandy2018}        & 29.63 & 29.03 \\
        \hline
        Mined forward   & 31.11 & 28.70 \\ 
        Mined backward  & {\bf 31.12}\makebox[0pt][l]{*} & {\bf 29.21}\makebox[0pt][l]{*} \\ 
        \hline
    \end{tabular}
    \caption{BLEU scores on WMT testing sets of the NMT models trained on original UN pairs (Oracle) and on two versions of mined UN corpora at sentence level. (*) indicates $p<0.001$ compared to the model trained on oracle pairs.}
    \label{tab:un_wmt}
\end{table}

\subsection{Evaluation Using a Translation Model}

After the retrieval step, we use cosine similarity scores to filter sentence pairs.
We empirically set the threshold for the cosine similarity to 0.5, based on the hold-out development set.
Then, we train NMT models on the filtered sentence pairs for en-es and en-fr.
The trained models are evaluated on the wmt13~\cite{wmt13} and wmt14~\cite{wmt14} test sets for en-es and en-fr, respectively.

To train the translation models, the sentence pairs are segmented using a shared 32,000 word-piece vocabulary \cite{schuster2012}.
We employ a 6-layer transformer architecture~\cite{vaswani2017attention} with a model dimension of 512, a hidden dimension of 2048, and 8 attention heads.
The Adam optimization algorithm is used with the training schedule described in \cite{vaswani2017attention}.
To ensure a fair comparison, our training setup is similar to the one used in \cite{mandy2018}, where sentences are batched by approximate sequence length with 128 sentences per batch.
We test the performance of this setup using the UN Oracle training set and confirm that our results are within 0.2 BLEU of the original numbers reported.

We evaluate the sentence pairs mined from the forward and backward directions separately.
Table \ref{tab:un_wmt} shows the BLEU~\cite{bleu} of the NMT models trained from the mined sentence pairs.
The first two rows show baseline models trained with the original UN pairs (Oracle) and from \cite{mandy2018} respectively.
All models perform very similarly, scoring within 2 BLEU points of each other.
The one trained with pairs mined from the backward search achieves the highest performance in both language pairs.
It is surprising to see that the model trained on mined data is even better than the model trained on original data.
To investigate why, we examined some \textit{``false positive''} examples returned by the nearest neighbor search.
We found that many of them actually are quite good translations, with only a few words different from the actual translation pairs,
some of them are even better aligned with the original translation. 
Our model is apparently filtering out some noise in the original data.
While our results show that raw cosine similarity can perform as well on the UN corpus as a classifier based on methods from prior work \cite{mandy2018}, we are not claiming that filtering with a more sophisticate classifier is not useful. Section 5 explores filtering using a BERT-based model \cite{bert}.

\begin{table}
\small
    \centering
    \begin{tabular}{| l | r r r r|}
    \hline
    Model & en-fr & en-es & en-ru & en-zh \\ \hline
    \cite{mandy2018}    & 89.0 & 90.5& -- & -- \\
    \cite{jakob2010}    & 93.4 & 94.4 & -- & -- \\ \hline
    Forward(Avg)         & 96.7 & 97.3 & {\bf 98.6} & 97.3 \\ 
    Backward(Avg)        & {\bf 96.9} & {\bf 97.8} & 98.2 & {\bf 97.4} \\ \hline
    \end{tabular}
    \caption{P@1 (\%) of target document retrieval on the UN corpus. Models attempt to select the true target from the entire corpus (85k documents). We simply take the average of the sentence embeddings from BiDE+AM models as the document embedding.}
    \label{tab:target_UN_DOC}
\end{table}

\subsection{Document-Level Retrieval}

We experiment with the proposed model at document-level. 
For each document, we first compute the embeddings of all its sentences using the sentence-level model, e.g. BiDE+AM.
Then, we average of all the sentence embeddings to obtain a document-level embedding. The document embeddings are used to perform approximate nearest neighbor search to identify document translations within the UN corpus.

Table \ref{tab:target_UN_DOC} shows the P@1 results at document-level.
The results of two baseline models are shown in rows 1 and 2. 
The baseline models require either a post-document matching step after the sentence-level retrieval~\cite{mandy2018} or a complex engineered system~\cite{jakob2010}.
Rows 3 and 4 provide the results for both forward and backward retrieval using document embeddings obtained by averaging BiDE+AM sentence embeddings.
Both en-fr and en-es models establish new state-of-the-art performance at 97\% P@1.
We also list the results of the averaging models for en-ru and en-zh.
It is worth noticing that the performance is consistent across all language pairs.

\begin{table*}[!htb]
\small
    \centering
    \begin{tabular}{l | l | r r r || r r r || r r r || r r r }
        \hline
        \multirow{2}{*}{Models} & \multirow{2}{*}{Direction} &\multicolumn{3}{c||}{fr-en} & \multicolumn{3}{c||}{de-en} & \multicolumn{3}{c||}{ru-en} & \multicolumn{3}{c}{zh-en} \\
        \cline{3-14}
        & &  P & R & F &  P & R & F &  P & R & F &  P & R & F\\ \hline
        \multirow{2}{*}{\cite{artetxe2018a}}
        & Forward  & 82.1 & 74.2 & 78.0 & 78.9 & 75.1 & 77.0 & - & - & - & - & - & - \\
        & Backward & 77.2 & 72.7 & 74.7 & 79.0 & 73.1 & 75.9 & - & - & - & - & - & - \\ \hline
        \multirow{2}{*}{BiDE+AM}
        & Forward  & {\bf 86.7} & {\bf 85.6} & {\bf 86.1} & {\bf 90.3} & {\bf 88.0} & {\bf 89.2} & {\bf 84.6} & {\bf 91.1} & {\bf 87.7} &  86.7 & {\bf 90.9} & {\bf 88.8} \\ 
        & Backward & 83.8 & 85.5 & 84.6 & 89.3 & 87.7 & 88.5 & 83.6 & 90.5 & 86.9 & {\bf 88.7} & 87.5 & 88.1\\ \hline
    \end{tabular}
    \caption{[P]recision, [R]ecall and [F]-score of BUCC training set with cosine similarity score used to optimize the threshold for best F score. Following the naming of \protect\cite{bucc2018}, we treat en as the target language and the other language as source in forward search. Backward is vice versa. We show the cosine similarity performance of \protect\cite{artetxe2018a} as baseline.}
    \label{tab:bucc_train}
\end{table*}

\subsection{Margin Value}

\begin{figure}
  \centering
  \includegraphics[width=.45\textwidth]{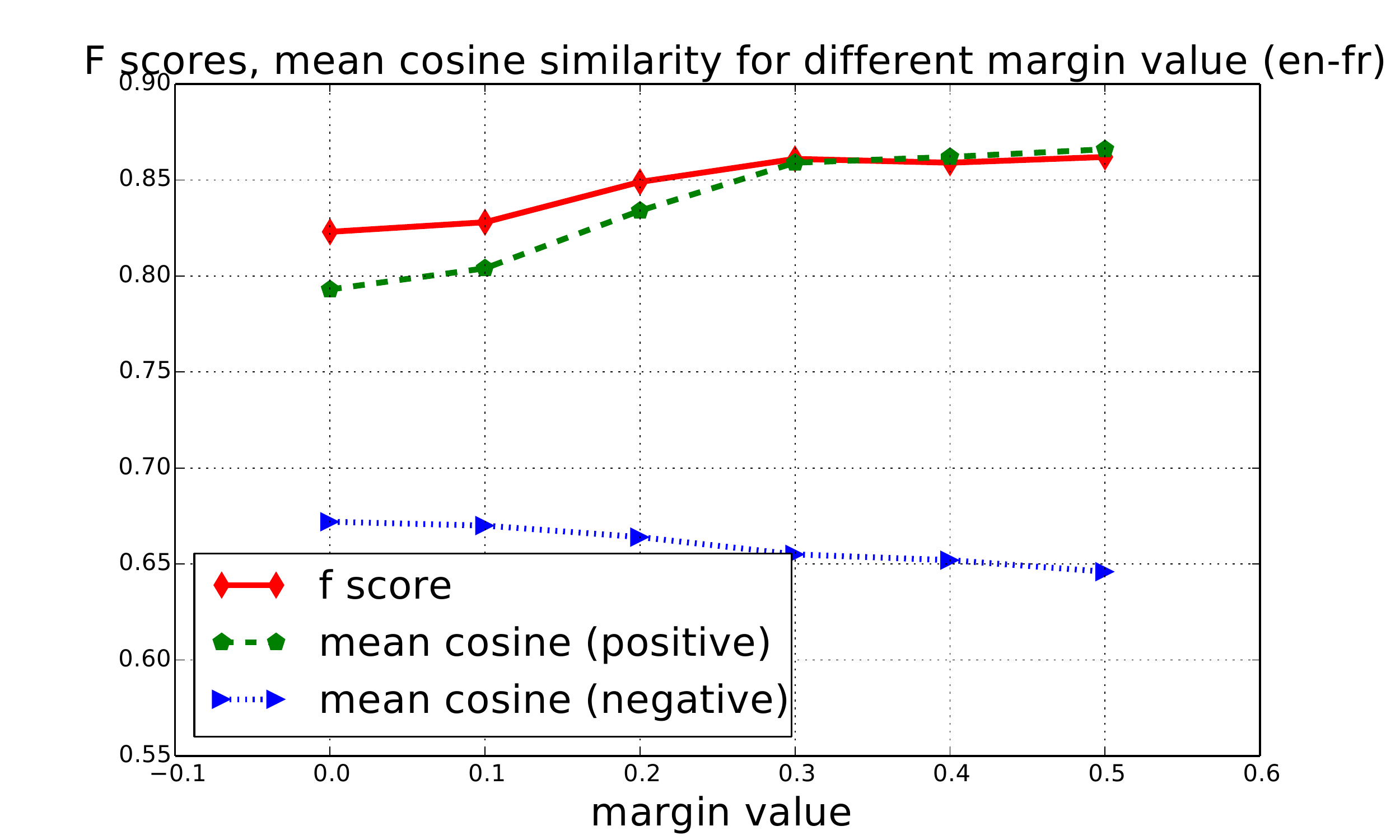}
  \caption{en-fr model performance and mean cosine similarity at different margin values.}
  \label{fig:margin_vs_no_margin}
\end{figure}

Figure \ref{fig:margin_vs_no_margin} shows the F1 scores for the en-fr model with different margin values, from 0 to 0.5. 
The model improved even with a small margin value, and it kept improving as the margin reached a value of 0.3.
The figure also shows the mean cosine similarity scores for both positive and negative pairs. We leveraged our retrieval system to generate negative pairs by using as target in the pair the sentence that was retrieved in second place when the correct target was retrieved in first place. This makes the negative pairs especially challenging to be distinguished from the positive ones.
The mean scores of positive pairs kept increasing with the margin value, while the mean of the negative pairs kept decreasing, showing the effectiveness of the additive margin softmax.

\section{Evaluation on the BUCC Mining Task}

In this section, we evaluate the proposed models on the BUCC mining task~\cite{bucc2018}. The
BUCC mining task is a shared task on parallel sentence extraction from two monolingual corpora with a subset of them assumed to be parallel, and that has been available since 2016.
We make use of the data from the 2018 shared task, which consists of corpora for four language pairs: fr-en, de-en, ru-en and zh-en.
For each language pair, the shared task provides a monolingual corpus for each language and a gold mapping list containing true translation pairs. These pairs are the ground truth.
The task is to construct a list of translation pairs from the monolingual corpora. The constructed list is compared to the ground truth, and evaluated in terms of the F1 measure.
For more details on this task refer to \cite{bucc2018}.

\begin{table}[]
\small
    \centering
    \begin{tabular}{|l|l|r|r|r|r|} \hline
        
        & Model   & fr-en & de-en & ru-en & zh-en  \\ \hline
        \multirow{3}{*}{Train.} 
        & Raw cosine                      &  86.14 &  89.22 &  87.69 &  88.81 \\
        & $\text{Scorer}_{Eq. \protect\ref{eq:margin_scorer}}$     &  {\bf 89.99} &  {\bf 92.63} &  {\bf 90.09} &  {\bf 92.48}  \\
        & $\text{Scorer}_{Eq. \protect\ref{eq:margin_scorer_2}}$    &  89.43 &  91.59 &  89.33 &  91.24  \\ 
        \hline
        \hline
        \multirow{5}{*}{Test} 
        & $\text{Scorer}_{Eq. \protect\ref{eq:margin_scorer}}$    & 90.02 & 92.26 & 89.18 & 92.81   \\ 
        & $\text{Scorer}_{Eq. \protect\ref{eq:margin_scorer_2}}$      & 89.56 & 91.42 & 88.31 & 91.89 \\
        & $\text{Scorer}_{BERT}$                         & {\bf 96.96} & {\bf 97.24} & {\bf 93.38} & {\bf 96.00}  \\ 
        \hhline{|~|=====|}
        & Baseline 1                      &  81.00 & 86.00 & 81.00 & 77.00 \\
        & Baseline 2                      &  92.89 & 95.58 & 92.57 & 92.03 \\
        \hline
    \end{tabular}
    \caption{The final F1 scores of BUCC shared task with rescoring. Baseline 1: best results from \protect\cite{bucc2018}. Baseline 2: \protect\cite{artetxe2018a} with margin based scorer.}
    \label{tab:bucc_final}
\end{table}

\begin{CJK*}{UTF8}{gbsn}
\begin{table*}[!htb]
    \small
    \centering
    \begin{tabular}{|c | l|}
        \hline
        \multicolumn{2}{|c|}{\bf False positives (cosine similarity = 0.96, 0.89)} \\
        en-000034142 & The Declaration of Brussels (1874) stated that the "honours and rights of the family...should be respected". \\
        zh-000090005 
        & ''布鲁塞尔宣言（1874年）表示，''家庭荣誉和权利…应当受到尊重. \\ 
        & (''Brussels Declaration (1874) stated that ''family honour and rights... should be respected.) \\ \hline
        
        en-000069836 & Males had a median income of 26,397 versus 25,521 for females. \\
        zh-000078235 
        & 男性人均年收入为 52,454，女性为43,750。 \\
        & (Male per capita income is 52,454 and female is 43,750.) \\ \hline
        
        \hline
        \multicolumn{2}{|c|}{\bf False negatives (cosine similarity = 0.74, 0.69)} \\
        en-000038309 & This is especially true of Brazil, the source of much of this mistrust. \\
        zh-000044511 
        & 其中巴西尤其如此，这个国家是南美国家联盟间不信任感的主要源头。 \\ 
        & (This is especially true of Brazil, which is the main source of distrust between the South American countries.) \\ \hline
        
        en-000027659 & But this may change as the source countries become richer and undergo rapid declines in birth rates. \\
	    zh-000023675 
	    & 但随着移民母国逐渐发达、出生率逐渐下降，美国的情况也会很快有所改变。\\
	    & (However, as the immigration mother country gradually develops and the birth rate gradually declines, \\
	    & the situation in the United States will soon change.) \\
        \hline
\end{tabular}
\caption{Failure cases drawn from the BUCC zh-en training set using cosine similarity scoring with a threshold of 0.78 tuned to maximize training set f-score. False positives are mined pairs that are not in the gold pairs. False negatives are gold pairs missed by the retrieval system or with scores lower than the threshold. The text in parentheses is the zh text translated using Google Translate. English glosses are provided for the Chinese sentences. We note that a number of errors are due to noisy pairs in the BUCC data.}
\label{tab:bucc_examples}
\end{table*}
\end{CJK*}

For each language pair, we iteratively retrieve the nearest neighbors for each source sentence.
We then filter the retrieved nearest-neighbor pairs using the cosine similarity scores.
Table \ref{tab:bucc_train} reports the precision, recall and F-score on the training set for both the forward and backward search.
The BUCC forward search identifies English translations of source sentences from another language.\footnote{The BUCC forward task contrasts with the UN forward task that identifies non-English translations of English sentences} 
The cosine score threshold is optimized by itself for comparison with \cite{artetxe2018a}, which is listed in rows 1 and 2.
Rows 3 and 4 show the performance of our model.
Both the forward and backward search achieve very good performance on all metrics, with F-score ranges from 84.6 to 89.2.
Both results perform better than the baseline models for fr-en and de-en. We obtain a large F-score performance gain of around 10 points.
Once again, the search from non-English to English (the forward direction in this case) works better than starting from English to retrieve the other languages.
The performance level is surprisingly stable across languages even for distinct language pairs such as en-zh.

\subsection{Margin Rescoring}

We experiment with using a margin-based second-stage scorer similar to the one proposed by~\cite{artetxe2018a}. Pairs are rescored using the formula: $s(x,y)=\phi(x,y) / margin(x,y) + \phi(x,y)$,
where  $\phi$ is the cosine similarity function and
\begin{equation}
\label{eq:margin_scorer}
    margin(x, y) = {\sum_{z \in NN_k(x)}\frac{\phi(x, z)}{2k}} + {\sum_{z \in NN_k(y)} \frac{\phi(z, y)}{2k}}
\end{equation}
The formula is slightly modified from~\cite{artetxe2018a} by adding the cosine similarity $\phi$ as part of the final score.\footnote{During early experiments, we found the new formulation consistently performed better than the original.} We also experiment with a simplified margin only considering one direction, which greatly reduces computation time and simplifies the system: 
\begin{equation}
\label{eq:margin_scorer_2}
    margin(x, y) = {\sum_{y=1}^{k} \frac{\phi(x, y)}{k}}
\end{equation}

\subsection{BERT Rescoring}

We explore using a rescoring classifier based on fine-tuning a multilingual BERT model~\cite{bert}.
For training, we use the nearest neighbor pairs mined from the BUCC training set.
Pairs in the gold map list are treated as positives, and the rest as negatives.
We sample the negative pairs so that the positive/negative ratio is 1:10.
80\% of them are used for training and 20\% are used for development. 
On the test set, we first retrieve the nearest neighbors and remove those pairs that are not nearest neighbor of each other.
Then, the fine-tuned BERT classifier is applied to select the positive pairs.

The final results on the BUCC task are shown in table \ref{tab:bucc_final}.
Margin-based rescoring still helps to improve the F1 scores 2 to 3 points on average, even using the simplified one-directional version from Eq. \ref{eq:margin_scorer_2}.
The BERT rescoring model significantly outperforms the current state-of-the-art models,  with F-scores ranging from 93.38 (ru-en) to 97.24 (de-en).

\begin{figure}
  \centering
  \includegraphics[width=.49\textwidth]{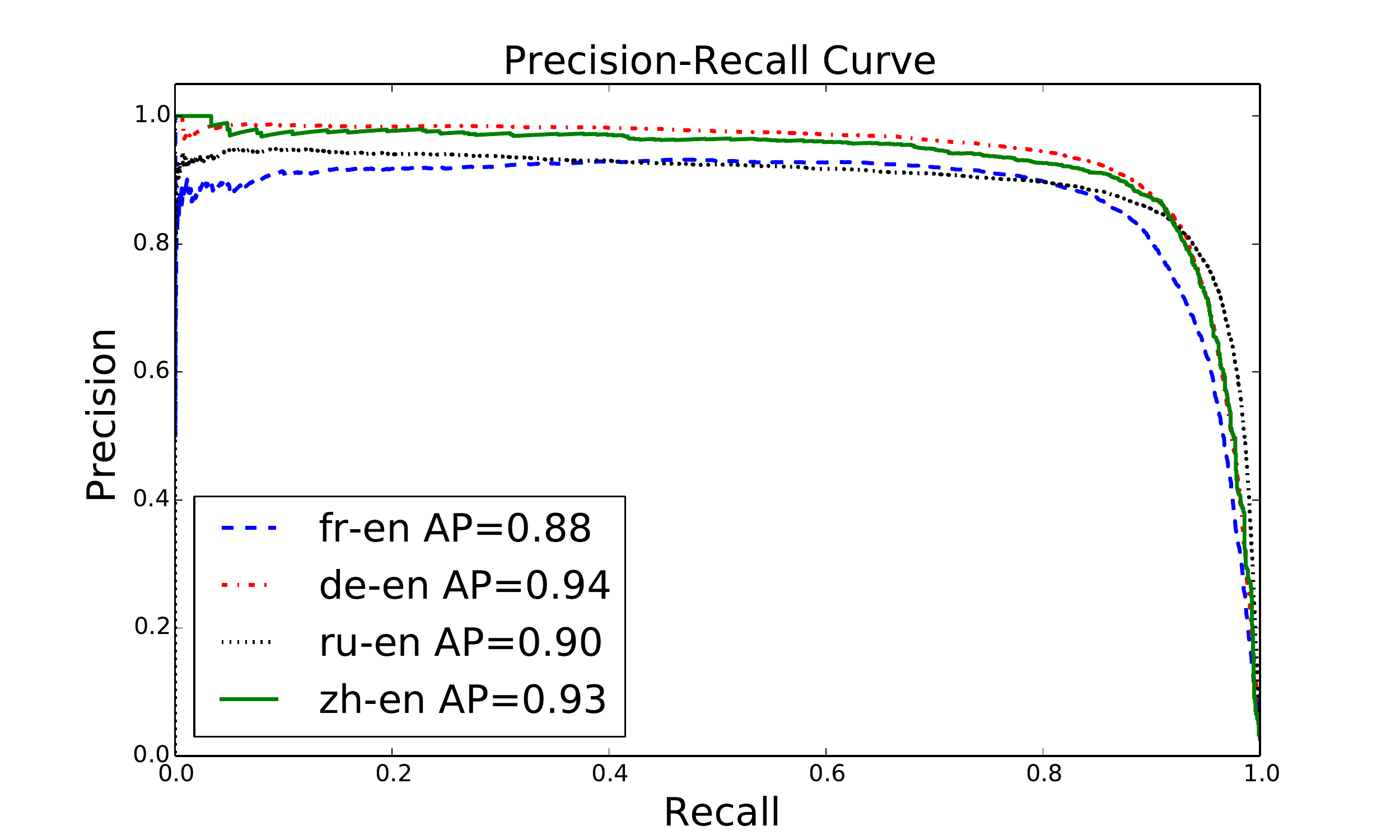}
  \caption{The precision-recall curve on the training set of the BUCC shared task, using a forward search with cosine similarly scoring. AP stands for ``Average Precision".}
  \label{fig:pr}
\end{figure}

\subsection{Analysis}
Figure \ref{fig:pr} shows the precision-recall curve on the training set of the BUCC shared task using a forward search with cosine similarly scoring.
The curves show that the similarity scores are quite stable and a strong translation quality signal for all language pairs. The figure also lists the average precision for each language pair, ranging from 88\% for fr-en to 94\% for de-en, which are also very promising.
    
To better understand the learned embedding space and what could be improved by a second-stage scorer,
we list typical failure cases drawn from the training set using a raw cosine similarity model for the zh-en task in table \ref{tab:bucc_examples}.
The first ``false positive'' is actually not really a failure. It is a good translation but that happens to be missing in the gold pairs.
\cite{bucc2018} mentioned that there are some true translations that are missing, but that they were not known to have affected the evaluation.
However, our models do seem to be identifying some true translations that are not accounted for by the BUCC data.
The second false positive is a typical error of the proposed model.
The model has trouble differentiating sentences that are semantically similar but that include different numbers or entities, especially  numbers with long digit sequences.
Part of the error may originate from the tokenization and normalization steps.
Interestingly, often several sentences are very close and with only differences in the numbers, e.g. \textit{(en-000065918) Males had a median income of \$59,738 versus \$39,692 for females; (en-000069337) Males had a median income of \$31,106 versus \$21,985 for females.}, etc.
As the pattern is very clear, they can be simply removed by a second-stage scorer.

We also list two typical false negatives from the gold pairs.
They are both ``partial translation'' pairs with cosine similarity close to the threshold.
These cases are especially challenging, as it is even hard for humans to make a consistent judgment on them. 
We found that some partial translations are present in the gold set, but others are not in the BUCC training task.
In practice, the threshold could be changed and fine-tuned based on the application use case.
This type of error seems hard to be captured by a margin-based scorer.
The BERT scorer, however, is likely capturing the human preference on the partial translations from the training data.
      
\section{Conclusion}
We present an approach to learn multilingual sentence embeddings using a bi-directional dual-encoder with additive margin softmax.
The resulting embedding space leads to better retrieval performance on the United Nations (UN) parallel corpus and on the BUCC task.
NMT systems trained on mined UN data retrieved using our models perform as well as NMT systems trained on the original UN bitext.
Document-level embeddings obtained by simply averaging the sentence level embeddings from our model achieve a new state-of-the-art on UN document-level mining. Our model achieves performance competitive with the current state-of-the art on BUCC.
When combined with a BERT rescoring model, our performance on BUCC achieves a new state-of-the-art.

\section*{Acknowledgments}
The authors are especially grateful to Wei Wang and Keith Stevens of the Google Translate Team for their valuable discussion and help running the data extraction and selection pipeline.

\bibliographystyle{named}
\bibliography{ijcai19}

\begin{thebibliography}{}

\bibitem[\protect\citeauthoryear{Antonova and Misyurev}{2011}]{antonova2011}
Alexandra Antonova and Alexey Misyurev.
\newblock Building a web-based parallel corpus and filtering out
  machine-translated text.
\newblock In {\em Proceedings of the 4th Workshop on Building and Using
  Comparable Corpora: Comparable Corpora and the Web}, pages 136--144.
  Association for Computational Linguistics, 2011.

\bibitem[\protect\citeauthoryear{Artetxe and Schwenk}{2018}]{artetxe2018a}
Mikel Artetxe and Holger Schwenk.
\newblock Margin-based parallel corpus mining with multilingual sentence
  embeddings.
\newblock {\em CoRR}, abs/1811.01136, 2018.

\bibitem[\protect\citeauthoryear{Bojar \bgroup \em et al.\egroup
  }{2013}]{wmt13}
Ond\v{r}ej Bojar, Christian Buck, Chris Callison-Burch, Christian Federmann,
  Barry Haddow, Philipp Koehn, Christof Monz, Matt Post, Radu Soricut, and
  Lucia Specia.
\newblock Findings of the 2013 {Workshop on Statistical Machine Translation}.
\newblock In {\em Proceedings of the Eighth Workshop on Statistical Machine
  Translation}, August 2013.

\bibitem[\protect\citeauthoryear{Bojar \bgroup \em et al.\egroup
  }{2014}]{wmt14}
Ond\v{r}ej Bojar, Christian Buck, Christian Federmann, Barry Haddow, Philipp
  Koehn, Johannes Leveling, Christof Monz, Pavel Pecina, Matt Post, Herve
  Saint-Amand, et~al.
\newblock Findings of the 2014 workshop on statistical machine translation.
\newblock In {\em Proceedings of the ninth workshop on statistical machine
  translation}, pages 12--58, 2014.

\bibitem[\protect\citeauthoryear{Bouamor and Sajjad}{2018}]{BOUAMOR18}
Houda Bouamor and Hassan Sajjad.
\newblock H2@bucc18: Parallel sentence extraction from comparable corpora using
  multilingual sentence embeddings.
\newblock In Reinhard Rapp, Pierre Zweigenbaum, and Serge Sharoff, editors,
  {\em Proceedings of the Eleventh International Conference on Language
  Resources and Evaluation (LREC 2018)}, may 2018.

\bibitem[\protect\citeauthoryear{Chen and Nie}{2000}]{chen2000parallel}
Jiang Chen and Jian-Yun Nie.
\newblock Parallel web text mining for cross-language ir.
\newblock In {\em Content-Based Multimedia Information Access-Volume 1}, pages
  62--77. Centre de Hautes Etudes Internationale D'Informatique Documentaire,
  2000.

\bibitem[\protect\citeauthoryear{Chidambaram \bgroup \em et al.\egroup
  }{2018}]{mutty2018}
Muthuraman Chidambaram, Yinfei Yang, Daniel Cer, Steve Yuan, Yun{-}Hsuan Sung,
  Brian Strope, and Ray Kurzweil.
\newblock Learning cross-lingual sentence representations via a multi-task
  dual-encoder model.
\newblock {\em CoRR}, abs/1810.12836, 2018.

\bibitem[\protect\citeauthoryear{Devlin \bgroup \em et al.\egroup
  }{2018}]{bert}
Jacob Devlin, Ming{-}Wei Chang, Kenton Lee, and Kristina Toutanova.
\newblock {BERT:} pre-training of deep bidirectional transformers for language
  understanding.
\newblock {\em CoRR}, abs/1810.04805, 2018.

\bibitem[\protect\citeauthoryear{Do \bgroup \em et al.\egroup
  }{2009}]{do2009mining}
Thi-Ngoc-Diep Do, Viet-Bac Le, Brigitte Bigi, Laurent Besacier, and Eric
  Castelli.
\newblock Mining a comparable text corpus for a vietnamese-french statistical
  machine translation system.
\newblock In {\em Proceedings of the Fourth Workshop on Statistical Machine
  Translation}, pages 165--172. Association for Computational Linguistics,
  2009.

\bibitem[\protect\citeauthoryear{dos Santos \bgroup \em et al.\egroup
  }{2016}]{attentive_pooling}
C{\'{\i}}cero~Nogueira dos Santos, Ming Tan, Bing Xiang, and Bowen Zhou.
\newblock Attentive pooling networks.
\newblock {\em CoRR}, abs/1602.03609, 2016.

\bibitem[\protect\citeauthoryear{Gr{\'e}goire and
  Langlais}{2017}]{gregoire2017}
Francis Gr{\'e}goire and Philippe Langlais.
\newblock A deep neural network approach to parallel sentence extraction.
\newblock {\em arXiv preprint arXiv:1709.09783}, 2017.

\bibitem[\protect\citeauthoryear{Guo \bgroup \em et al.\egroup
  }{2018}]{mandy2018}
Mandy Guo, Qinlan Shen, Yinfei Yang, Heming Ge, Daniel Cer, Gustavo
  Hernandez~Abrego, Keith Stevens, Noah Constant, Yun-hsuan Sung, Brian Strope,
  and Ray Kurzweil.
\newblock Effective parallel corpus mining using bilingual sentence embeddings.
\newblock In {\em Proceedings of the Third Conference on Machine Translation:
  Research Papers}, pages 165--176. Association for Computational Linguistics,
  2018.

\bibitem[\protect\citeauthoryear{Hassan \bgroup \em et al.\egroup
  }{2018}]{hassan2018achieving}
Hany Hassan, Anthony Aue, Chang Chen, Vishal Chowdhary, Jonathan Clark,
  Christian Federmann, Xuedong Huang, Marcin Junczys-Dowmunt, William Lewis,
  Mu~Li, et~al.
\newblock Achieving human parity on automatic chinese to english news
  translation.
\newblock {\em arXiv preprint arXiv:1803.05567}, 2018.

\bibitem[\protect\citeauthoryear{Munteanu and
  Marcu}{2005}]{munteanu2005improving}
Dragos~Stefan Munteanu and Daniel Marcu.
\newblock Improving machine translation performance by exploiting non-parallel
  corpora.
\newblock {\em Computational Linguistics}, 31(4):477--504, 2005.

\bibitem[\protect\citeauthoryear{Munteanu and
  Marcu}{2006}]{munteanu2006extracting}
Dragos~Stefan Munteanu and Daniel Marcu.
\newblock Extracting parallel sub-sentential fragments from non-parallel
  corpora.
\newblock In {\em Proceedings of the 21st International Conference on
  Computational Linguistics and the 44th Annual Meeting of the Association for
  Computational Linguistics}, pages 81--88. Association for Computational
  Linguistics, 2006.

\bibitem[\protect\citeauthoryear{Papineni \bgroup \em et al.\egroup
  }{2002}]{bleu}
Kishore Papineni, Salim Roukos, Todd Ward, and Wei-Jing Zhu.
\newblock Bleu: a method for automatic evaluation of machine translation.
\newblock In {\em Proceedings of the 40th annual meeting on association for
  computational linguistics}, pages 311--318. Association for Computational
  Linguistics, 2002.

\bibitem[\protect\citeauthoryear{Resnik and Smith}{2003}]{resnik2003web}
Philip Resnik and Noah~A Smith.
\newblock The web as a parallel corpus.
\newblock {\em Computational Linguistics}, 29(3):349--380, 2003.

\bibitem[\protect\citeauthoryear{Resnik}{1999}]{resnik1999mining}
Philip Resnik.
\newblock Mining the web for bilingual text.
\newblock In {\em Proceedings of the 37th Annual Meeting of the Association for
  Computational Linguistics on Computational Linguistics}, pages 527--534.
  Association for Computational Linguistics, 1999.

\bibitem[\protect\citeauthoryear{Schuster and Nakajima}{2012}]{schuster2012}
M.~Schuster and K.~Nakajima.
\newblock Japanese and korean voice search.
\newblock In {\em 2012 IEEE International Conference on Acoustics, Speech and
  Signal Processing (ICASSP)}, pages 5149--5152, March 2012.

\bibitem[\protect\citeauthoryear{Schwenk}{2018}]{schwenk2018filtering}
Holger Schwenk.
\newblock Filtering and mining parallel data in a joint multilingual space.
\newblock {\em arXiv preprint arXiv:1805.09822}, 2018.

\bibitem[\protect\citeauthoryear{Shi \bgroup \em et al.\egroup
  }{2006}]{shi2006dom}
Lei Shi, Cheng Niu, Ming Zhou, and Jianfeng Gao.
\newblock A dom tree alignment model for mining parallel data from the web.
\newblock In {\em Proceedings of the 21st International Conference on
  Computational Linguistics and the 44th Annual Meeting of the Association for
  Computational Linguistics}, pages 489--496. Association for Computational
  Linguistics, 2006.

\bibitem[\protect\citeauthoryear{Uszkoreit \bgroup \em et al.\egroup
  }{2010}]{jakob2010}
Jakob Uszkoreit, Jay~M. Ponte, Ashok~C. Popat, and Moshe Dubiner.
\newblock Large scale parallel document mining for machine translation.
\newblock In {\em Proceedings of the 23rd International Conference on
  Computational Linguistics}, COLING '10, 2010.

\bibitem[\protect\citeauthoryear{Utiyama and
  Isahara}{2003}]{utiyama2003reliable}
Masao Utiyama and Hitoshi Isahara.
\newblock Reliable measures for aligning japanese-english news articles and
  sentences.
\newblock In {\em Proceedings of the 41st Annual Meeting on Association for
  Computational Linguistics}, pages 72--79. Association for Computational
  Linguistics, 2003.

\bibitem[\protect\citeauthoryear{Vanderkam \bgroup \em et al.\egroup
  }{2013}]{ann}
Dan Vanderkam, Rob Schonberger, Henry Rowley, and Sanjiv Kumar.
\newblock Nearest neighbor search in google correlate.
\newblock Technical report, Google, 2013.

\bibitem[\protect\citeauthoryear{Vaswani \bgroup \em et al.\egroup
  }{2017}]{vaswani2017attention}
Ashish Vaswani, Noam Shazeer, Niki Parmar, Jakob Uszkoreit, Llion Jones,
  Aidan~N Gomez, {\L}ukasz Kaiser, and Illia Polosukhin.
\newblock Attention is all you need.
\newblock In {\em Advances in Neural Information Processing Systems}, pages
  5998--6008, 2017.

\bibitem[\protect\citeauthoryear{Wang \bgroup \em et al.\egroup }{2018a}]{ams}
Feng Wang, Jian Cheng, Weiyang Liu, and Haijun Liu.
\newblock Additive margin softmax for face verification.
\newblock {\em IEEE Signal Processing Letters}, 25(7):926--930, 2018.

\bibitem[\protect\citeauthoryear{Wang \bgroup \em et al.\egroup
  }{2018b}]{wei2018}
Wei Wang, Taro Watanabe, Macduff Hughes, Tetsuji Nakagawa, and Ciprian Chelba.
\newblock Denoising neural machine translation training with trusted data and
  online data selection.
\newblock In {\em Proceedings of the Third Conference on Machine Translation},
  pages 133--143. Association for Computational Linguistics, 2018.

\bibitem[\protect\citeauthoryear{Yang and Li}{2002}]{yang2002mining}
Christopher~C Yang and Kar~Wing Li.
\newblock Mining english/chinese parallel documents from the world wide web.
\newblock In {\em Proceedings of the 11th International World Wide Web
  Conference, Honolulu, USA}, 2002.

\bibitem[\protect\citeauthoryear{Yang \bgroup \em et al.\egroup
  }{2018}]{yang2018}
Yinfei Yang, Steve Yuan, Daniel Cer, Sheng-Yi Kong, Noah Constant, Petr Pilar,
  Heming Ge, Yun-hsuan Sung, Brian Strope, and Ray Kurzweil.
\newblock Learning semantic textual similarity from conversations.
\newblock In {\em Proceedings of The Third Workshop on Representation Learning
  for NLP}. Association for Computational Linguistics, 2018.

\bibitem[\protect\citeauthoryear{Yin \bgroup \em et al.\egroup
  }{2015}]{yin2015abcnn}
Wenpeng Yin, Hinrich Sch{\"u}tze, Bing Xiang, and Bowen Zhou.
\newblock Abcnn: Attention-based convolutional neural network for modeling
  sentence pairs.
\newblock {\em arXiv preprint arXiv:1512.05193}, 2015.

\bibitem[\protect\citeauthoryear{Ziemski \bgroup \em et al.\egroup
  }{2016}]{uncorpus}
Michal Ziemski, Marcin Junczys-Dowmunt, and Bruno Pouliquen.
\newblock The united nations parallel corpus v1. 0.
\newblock April 2016.

\bibitem[\protect\citeauthoryear{Zweigenbaum \bgroup \em et al.\egroup
  }{2018}]{bucc2018}
Pierre Zweigenbaum, Serge Sharoff, and Reinhard Rapp.
\newblock Overview of the third bucc shared task: Spotting parallel sentences
  in comparable corpora.
\newblock In Reinhard Rapp, Pierre Zweigenbaum, and Serge Sharoff, editors,
  {\em Proceedings of the Eleventh International Conference on Language
  Resources and Evaluation (LREC 2018)}, may 2018.

\end{thebibliography}

\end{document}